\documentclass[twoside]{article}
\usepackage{ecj,palatino,epsfig,latexsym,natbib}

\usepackage{subfig}
\usepackage{placeins}
\usepackage{amsmath}
\usepackage{algorithmic}
\usepackage{array}

\usepackage{url}
\usepackage{rotating}
\usepackage[percent]{overpic}
\usepackage[dvipsnames]{xcolor}

\usepackage{contour}
\contournumber{64}
\contourlength{.015em}

\usepackage{tcolorbox}

\newenvironment{hypothesis}%
  {\list{}{\leftmargin=8.7mm\rightmargin=0.0mm}\item[]}%
  {\endlist}

\begin{document}

\ecjHeader{x}{x}{xxx-xxx}{2020}{Evolutionary Adaptation of Morphology and Control}{T.F. Nygaard, C.P. Martin, D. Howard, J. Torresen, K. Glette}
\title{\bf Environmental Adaptation of Robot Morphology and Control through\\ Real-world Evolution}  

\author{\name{\bf T. F. Nygaard} \hfill \addr{tonnesfn@ifi.uio.no}\\ 
        \addr{Department of Informatics, University of Oslo, Norway}\\
        \addr{Norwegian Defence Research Establishment, Kjeller, Norway}
\AND
        \name{\bf C. P. Martin} \hfill \addr{charles.martin@anu.edu.au}\\
        \addr{Research School of Computer Science, Australian National University, ACT, Australia}
\AND
        \name{\bf D. Howard} \hfill \addr{david.howard@data61.csiro.au}\\
        \addr{Cyber-Physical Systems Program, CSIRO, QLD, Australia}
\AND
        \name{\bf J. Torresen} \hfill \addr{jimtoer@ifi.uio.no}\\
        \addr{Department of Informatics, University of Oslo, Norway}\\
        \addr{RITMO, University of Oslo, Norway}
\AND
        \name{\bf K. Glette} \hfill \addr{kyrrehg@ifi.uio.no}\\
        \addr{Department of Informatics, University of Oslo, Norway}\\
        \addr{RITMO, University of Oslo, Norway}
}

\maketitle

\begin{abstract}
  Robots operating in the real world will experience a range of different environments and tasks.
  It is essential for the robot to have the ability to adapt to its surroundings to work efficiently in changing conditions.
  Evolutionary robotics aims to solve this by optimizing both the control and body (morphology) of a robot, allowing adaptation to internal, as well as external factors.
  Most work in this field has been done in physics simulators, which are relatively simple and not able to replicate the richness of interactions found in the real world.
  Solutions that rely on the complex interplay between control, body, and environment are therefore rarely found.
  In this paper, we rely solely on real-world evaluations and apply evolutionary search to yield combinations of morphology and control for our mechanically self-reconfiguring quadruped robot.
  We evolve solutions on two distinct physical surfaces and analyze the results in terms of both control and morphology.
  We then transition to two previously unseen surfaces to demonstrate the generality of our method.
  We find that the evolutionary search finds high-performing and diverse morphology-controller configurations by adapting both control and body to the different properties of the physical environments.  We additionally find that morphology and control vary with statistical significance between the environments.
  Moreover, we observe that our method allows for morphology and control parameters to transfer to previously-unseen terrains, demonstrating the generality of our approach.

\end{abstract}


\section{Introduction}
The evolutionary theory describes how animals exhibit behavioral, structural, and physiological adaptations to environmental changes across multiple generations, which increases the likelihood of survival and the preservation of their genes.
In nature, this process is dependent on a large number of generations of animals breeding and raising their young, resulting in many years for adaptation to take place.
Natural organisms can adapt through learning reasonably quickly, but their morphological adaptation is a long process. In robotics, we can also learn quickly through, e.g., controller adaptation, but as opposed to nature, we can also perform morphological adaptation in real-time. 

In the context of robotics, adaptability to dynamic environmental conditions and mission parameters is a key enabling technology that allows robots to perform increasingly complex tasks in challenging environments.
Practically, improving the adaptability of a robot unlocks an ever-expanding repertoire of deployment scenarios such as disaster response, autonomous surveying, and others. 

Legged robots are noted for their agility and ability to traverse a multitude of terrain types and, as such, hold a particular promise for completing such missions~\citep{Hwangboeaau5872}.
They also provide an intrinsic and straightforward route towards adaptability, in that their controllers, typically the rhythms produced by a gait engine or the arcs that describe the movement of the robot's foot-tip positions, can easily be optimized through, e.g., reinforcement learning~\citep{kohl2004policy} and evolutionary techniques ~\citep{heijnen2017testbed}.

\begin{figure}
  \centering
  \includegraphics[width=0.9\textwidth]{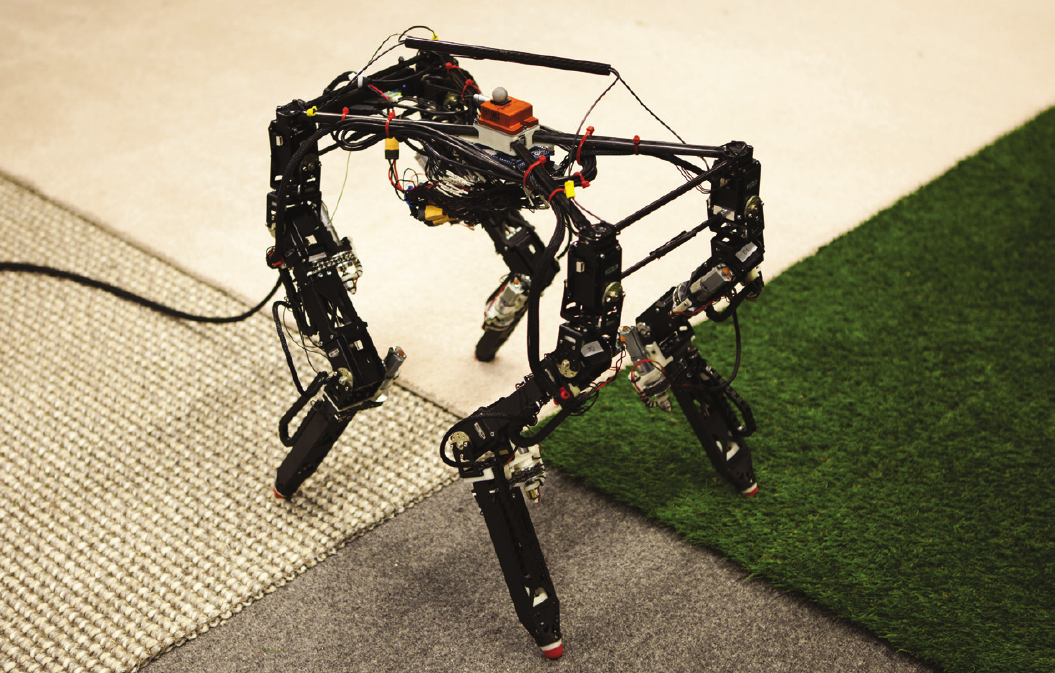} 
  \caption{The Dynamic Robot for Embodied Testing (DyRET) standing on the four different surfaces used in our experiments. The robot can change the length of its legs to adapt its body to the environment it operates in.}
  \label{fig.robot}
\end{figure}

As a simple motivating case, consider a legged robot deployed in a forest or garden.
The robot can reasonably be expected to have to scramble over some obstacles as well as squeeze between others.  
It follows that high in-mission performance requires the robot to be tall (at times) to step over obstacles, and small (at other times) to squeeze into tight spaces.
In cases like these, behavioral adaptation through pure controller optimization, as described above, provides only {\em limited adaptability}, in that the morphology of the robot is static\footnote{although posture may change through control, the actual geometry of the robot is fixed}.
Recent work has shown examples where morphological adaptation of a physical robot can serve as an effective alternative to classic adaptation of control~\citep{kriegman19_2}, and even cases where adapting morphology is the only feasible option~\citep{Giacomo19}.

We describe an approach that bridges the fields of embodied intelligence \citep{howard2019evolving} and evolvable hardware~\citep{evohw-book} by focusing on in-environment behaviors produced by the simultaneous evolution of controller and morphology.
All experiments are conducted in hardware on a dynamically reconfigurable quadruped platform (shown in Fig. \ref{fig.robot}).
Embodied AI tells us that intelligent in-environment behavior arises from a strong link between morphology, controller, and environment.
By co-evolving morphology and controller on a hardware-reconfigurable robot, we can expect to perform a broader range of missions in more challenging scenarios than if just controller tuning was considered~\citep{nygaard_WS_ICRA18}.
The relatively few previous studies on the simultaneous evolution of legged robot morphology and control have mostly performed the evolution in simulated environments \citep{tonnesfn_evostar17, Auerbach19, miras2019effects, hornby1999autonomous}.
The few examples performed on real-world systems are on simple robots or require too much human intervention or time to adapt to continuously changing environments \citep{vujovic2017evolutionary, milan17}.

Our earlier experiments on real-world evolution of legged robot morphology and control include adaptation to different \emph{internal} hardware states~\citep{tonnesfn_gecco18}.
In this paper, we tackle the challenge of real-world adaptation to \emph{external} states.
Specifically, we conduct experiments on adapting robot morphology and control to different types of planar surfaces, see Fig.~\ref{fig.robot}.

We propose two hypotheses to investigate robotic structural adaptation in the real world:
\begin{enumerate}
  \item[H1:] \emph{Performing an evolutionary search in diverse physical environments will result in individuals with significantly different control and morphology.}
\end{enumerate}
\begin{hypothesis}
  This states that the evolutionary search adapts the individuals to the environments they are evolved in and can be disproven if we observe quantitatively similar individuals after evolving on characteristically different terrains. 
\end{hypothesis}

\begin{enumerate}
    \item[H2:] \emph{The performance of the evolved individuals will transfer better to qualitatively similar environments.}
\end{enumerate}
\begin{hypothesis}
    This states that the results found in evolution will generalize and that individuals adapted to one type of terrain will perform comparably in other qualitatively similar terrains. It can be disproven if the performance of individuals is shown to be very different in similar terrains.
\end{hypothesis}
  
Our results reveal that evolving on different surfaces has a significant effect on both the control and the morphology of the robot. 
This supports our first hypothesis on adaptation.
Subsequently, we tested the resulting individuals from the evolutionary runs on previously unseen surfaces. 
The results show that individuals perform best on surfaces that are qualitatively close to the one they were trained on.
This supports our second hypothesis on generalization to unseen environments.

Being able to handle unknown, dynamic environments is a compelling reason for adaptation, and our ultimate goal.
In this paper, we contribute with an important first step by testing adaptation on a variety of indoor planar surfaces by showing that morphological adaptation is a crucial factor in unlocking heightened performance, even in these relatively innocuous environmental scenarios.
Moreover, in the vein of Auerbach and Bongard's work~\citep{auerbach2014environmental}, we show that evolution can adapt to different terrains, but this time using a real-world-only search. Adaptation to these kinds of environmental differences is highly dependent on real-world evaluations as the natural noise, uncertainty, and detailed interaction dynamics can be impossible to model with current physics simulators~\citep{grandchallenges}.

\section{Background}
In this section, we provide a brief overview of the field of evolutionary robotics, with a particular focus on gait learning, morphological adaptation, and optimization on physical robots in the real world.

Most effort in robot adaptation has focused on improving the control of a robot.
Typically, the gait pattern~\citep{weingarten2004automated}, foot-tip arcs~\citep{heijnen2017testbed}, or more high-level gait parameters~\citep{tonnesfn_ices16} are candidates for optimization.
Approaches include heuristic terrain adaptation~\citep{Homberger2016Terrain-dependant} to switch gait based on, e.g., energetics~\citep{Kottege2015Energetics} and power consumption~\citep{jin2013power} as terrain changes, as well as more complex hierarchical optimization-based approaches~\citep{7803330}.
There are many methods to optimize the control of a robot, including reinforcement learning~\citep{kohl2004policy}, transfer learning~\citep{degrave2015transfer}, and deep reinforcement learning~\citep{Hwangboeaau5872}.
Another popular approach is to use methods from evolutionary computation~\citep{whatwhywhere}.
Many different types of legged robots are used in research, but doing machine learning on physical robots puts high demands on reliability and maintainability, making some robots more suitable than others~\citep{nygaard2019experiences}.

Evolutionary robotics is a field that uses techniques from evolutionary computation to optimize different aspects of robots.
Most work in the field has traditionally been done on virtual robots in simplified physics simulations and is only concerned with optimizing the controller~\citep{mouret201720}.
Working in simplified simulations alone often results in controllers and morphologies that are hard to transfer to the real world due to inaccuracies in the modeling, a problem referred to as the \textit{reality gap}.
Many solutions to reduce the reality gap have been proposed, such as adding noise~\citep{jakobi1995noise}, starting in simulation and finishing in the real world~\citep{Nolfi:63867}, modeling the reality gap itself~\citep{koos2013transferability}, and treating simulators as just another environment that needs adaptation ~\citep{nordmoen19evolved}.
There have been several significant contributions to solve this problem, but they have not kept up with the increased complexity in the terrains and environments of current robots~\citep{mouret201720}.

There are many examples of evolutionary robotics techniques being used to evolve gaits in the real world on different physical robots, including commercial off-the-shelf legged robots like the AIBO ~\citep{hornby1999autonomous,chernova2004evolutionary}, as well as purpose-built custom robots~\citep{yosinski2011evolving}.
Most optimize a limited set of parameters that control all the legs identically.
However, there are also examples that generate separate control arcs for each leg~\citep{heijnen2017testbed}, which allows adaptation to the specifics of the hardware (e.g., an actuator slipping, or delivering reduced torque due to wear).

Optimizing control can be a very effective way of adapting a robot to new tasks and environments.
However, there are several examples where optimizing the body of the robot, its morphology, can also be an effective method.
Due to the inherent difficulties associated with making a variable-morphology robot, most research to date on adaptive morphology on physical robots does the optimization ahead of time in simulation, then transfers a select few individuals to the real world for testing.
Examples of this include legged robots~\citep{tonnesfn_evostar17}, soft robots \citep{kriegman19}, modular robots~\citep{Auerbach19}, and even flying robots~\citep{rosser19}.

\vspace{5mm}

Morphological adaptation offers more freedom to tailor in-environment behavior, and through the lens of embodied cognition, provides the means to tightly couple control and morphology with environmental performance. Morphological adaptation approaches in hardware are now becoming increasingly viable. 
This is largely thanks to ongoing improvements in the quality and availability of the prerequisite robotic components, and the rapid adoption of flexible fabrication techniques, e.g., 3D printing, into the robot design process.
There are broadly two approaches to achieve morphological robot adaptation: (i) optimize 3D-printable components and attach them to a robotic base to provide bespoke terrain-specific performance properties~\citep{collins2018towards}; or (ii) create a single robot with built-in adaptation abilities~\citep{tonnesfn19icra}.
We focus on the latter approach as it allows morphology to be changed in-situ, e.g., as a real-time response to environmental stimuli.
It also opens up the possibility to incrementally learn a controller on a simpler hardware configuration, which has previously been shown to increase robustness \citep{Bongard1234}.

We note that the literature shows a distinct lack of morphological variation on real-world robots, making our focus on employing evolutionary methods for optimization of the morphology of a physical quadruped robot novel.

\section{Materials and Methods}
This section introduces our robot, before describing the gait controller, evolutionary setup, and physical environments used in the experiments. 

\subsection{The DyRET Robot}

\begin{figure}
  \centering
  \includegraphics[width=0.81\textwidth]{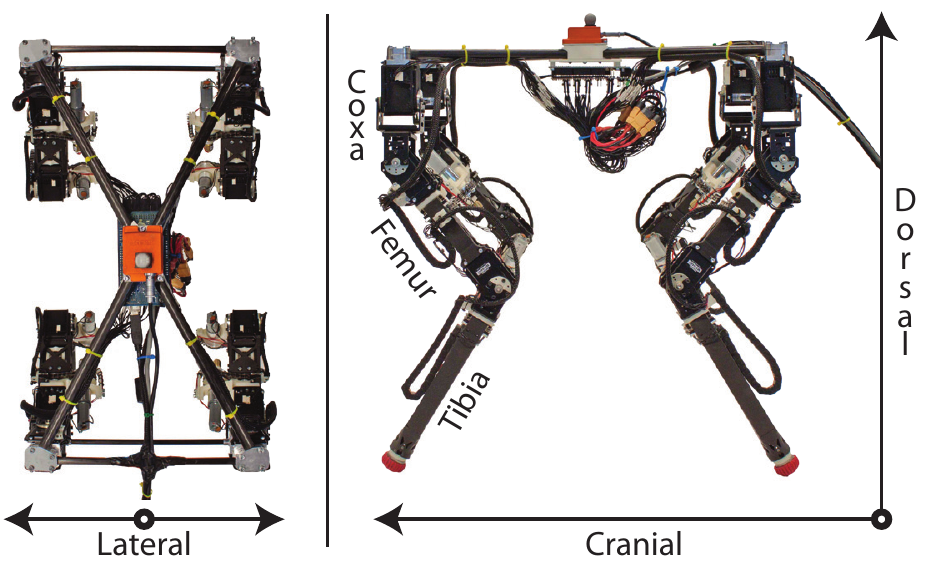}
  \caption{Top and side view of the robot, showing the directional terminology used as well as the names of the three links of the legs.}
  \label{fig.dyretAxes}
\end{figure}

We used the Dynamic Robot for Embodied Testing (DyRET), a mammal-inspired quadruped robot~\citep{tonnesfn19icra}.
This custom robot was developed at the University of Oslo as a platform for evolutionary experiments, in particular for simultaneous optimization of morphology and control.
The design allows the robot to independently and automatically change the length of its femurs and tibias, as seen in Fig.~\ref{fig.dyretAxes}. 

The DyRET project is certified by the Open Source Hardware Association (OSHWA) as fully open source.
All software, hardware design files, documentation, and simulation models are available online\footnote{https://github.com/dyret-robot/dyret\_documentation}.

The robot body is mostly built with 3D-printed fiber-reinforced plastic and milled aluminum parts, along with Commercial-Off-The-Shelf (COTS) parts, including carbon fiber tubing and aluminum brackets.
Being able to endure bad morphology and gait combinations can require a superabundance of motor power.
Keeping weight down was, for this reason, a priority during the design phase, while keeping high robustness and maintainability of the platform.

Each leg features three revolute joints, with a Dynamixel MX-64 in the coxa, and Dynamixel MX-106 servos in the femur and tibia.
These are connected on a common bus, and each run a separate PD position controller.
Each femur and tibia consist of a custom linear actuator that allows the length of each leg segment to be changed. 
The mechanical means of achieving repeatable, physically strong morphological reconfiguration on a robot of this size requires the use of, e.g., screw-based linear actuators, the trade-off being that the mechanism is too slow to be used actively as actuation in a dynamic gait.
The femur can extend by 50mm, and the tibia by 100mm.
This is powered by a brushed DC motor, connected to a lead screw through a chain, resulting in a linear speed of approximately 1mm/sec.
An absolute encoder gives the linear actuators an accuracy of around half a millimeter. 

\begin{figure}
  \centering
  \includegraphics[width=0.85\textwidth]{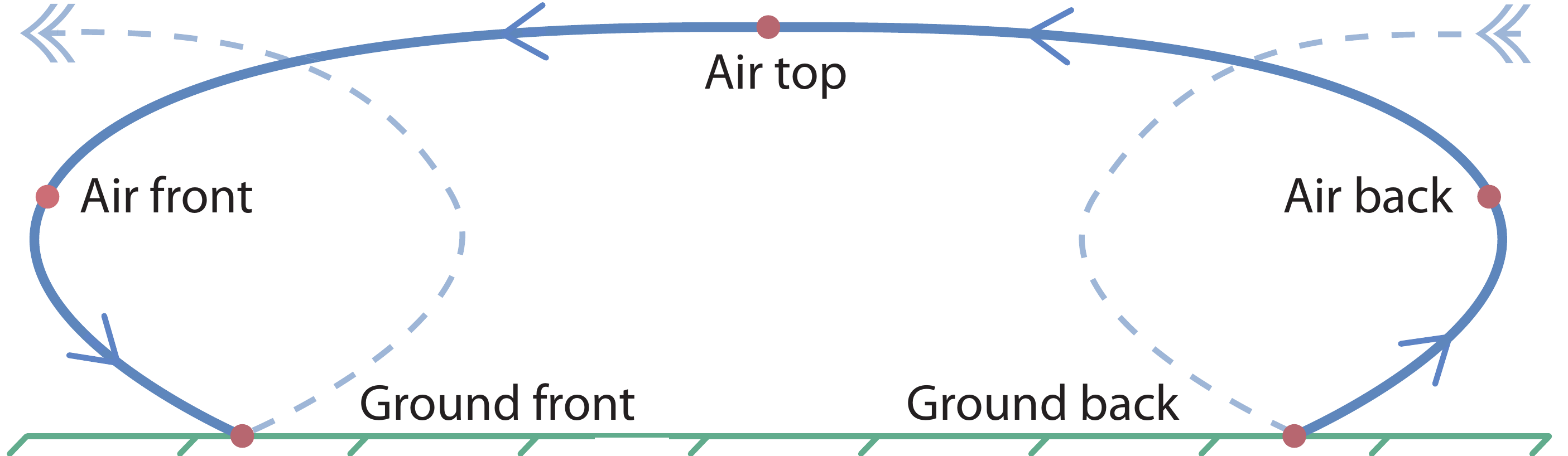}
  \caption{An example of a typical foot tip trajectory, including the five control points used to calculate the spline shape. The solid line shows the full step, while the dashed lines show the trajectory for the previous and next steps. This view shows the side of the robot, with the front of the robot to the left of the figure.}
  \label{fig.splineShape}
\end{figure}

The robot features an MTi-30 Attitude and Heading Reference System (AHRS) in the center of its body, which gives absolute orientation, rotational velocity, and linear acceleration.
Reflective markers for motion capture are mounted on the robot body for measuring absolute position and orientation.
Each servo has sensors for temperature, voltage, current, and joint position.

The robot is tethered to a desktop computer over USB, which runs all software on the Robot Operating System (ROS) framework.
The gait controller runs at a rate of 50Hz, exchanging commands and data to and from the servos, while the sensors and motion capture system are sampled at 100Hz.

\subsection{Parameterizable Control and Morphology}

The robot uses a high-level spline-based gait controller working in Cartesian space~\citep{nygaard_evostar19}.
For generating the foot tip trajectory, a three-dimensional looping cubic Hermite spline is built using five control points.
These define the path the leg takes through the air, as seen in Fig.~\ref{fig.splineShape}.
All legs follow the same trajectory but are offset with a static phase shift.
The adjustable parameters and ranges of the parameters can be seen in Table~\ref{table.splineParams}.
Note that the controller only generates target positions and that the actual leg movement can vary depending on the surface it is walking on.

In addition to the 11 parameters controlling the spline shape, seven global parameters are shown in Table~\ref{table.gaitParams}.
Two parameters define the gait timing; the frequency of the gait (steps per second for each leg), and lift duration (the percentage of the gait period spent moving the leg through the air).
We have also added a balancing wag---a counter-balancing movement---to increase the stability of the robot by leaning to the opposite side of the leg it is currently lifting. 
This movement is needed since the weight of each leg is high compared to the overall body weight.
It is especially useful at slow speeds, where the inertia from movement alone can not be used to counteract the shifting weight of the legs. 
The phase of this movement can be changed, along with separate amplitudes for the lateral and cranial amplitude (see Fig.~\ref{fig.dyretAxes}).

For controlling the morphology, the \emph{Femur length} and \emph{Tibia length} parameters are used. 
For the experiments in this paper, all legs share the same femur and tibia lengths. 
The legs will reconfigure fully to the target length before the evaluation of a new gait is performed.

\begin{table}
    \caption{Parameters and ranges defining the spline shape. The axes are shown in Fig.~\ref{fig.dyretAxes}.}
    \label{table.splineParams}
    \centering
    \begin{tabular}{ l | r r r}
        \hline
        Control point & Lateral & Cranial & Dorsal \\
        \hline
        Ground\textsubscript{front} & 0 & [0, 100] & 0  \\
        Ground\textsubscript{back}  & 0 & [-150, -50] & 0  \\
        Air\textsubscript{front} & [-12.5, 12.5] & [25, 125] & [19, 41] \\
        Air\textsubscript{top} & [-12.5, 12.5] & [-30, 30] & [39, 61]  \\
        Air\textsubscript{back} & [-12.5, 12.5] & [-125, -25] & [19, 41]  \\
        \hline
    \end{tabular}
\end{table}

\begin{table}
    \caption{Ranges for gait and morphology parameters.}
    \label{table.gaitParams}
    \centering
    \begin{tabular}{ l | r r r r }
        \hline
        Parameter & Range \\
        \hline
        Wag phase       & $\approx$[-0.394, 0.394] \\
        Wag amplitudes  & [0, 14.0] \\
        Lift duration   & [0.13, 0.20] \\
        Frequency       & [0.25, 1.0] \\
        Femur length    & [0, 50] \\
        Tibia length    & [0, 100] \\
        \hline
    \end{tabular}
\end{table} 

\subsection{Evolutionary Setup}

Our evolutionary search was configured to find combinations of control and morphology that achieve both high speed and stability, and expose the Pareto front of trade-offs between these two objectives. 
For this, we apply the commonly used NSGA-II~\citep{nsga} algorithm with two fitness objectives, $F_{speed}$ and $F_{stability}$.

Speed is calculated by using the position of the robot from motion capture equipment ($P$) and can be seen in Equation~\ref{eq.fitSpeed}.
 Stability is found using the on-board Attitude and Heading Reference System (AHRS) sensor, calculated as a combination between standard deviations of orientation ($Ang$) and linear acceleration ($Acc$), seen in Equation~\ref{eq.fitStab}.
For our sensor, $\alpha$ has experimentally been set to $\frac{1}{50}$, resulting in approximately equal contributions from linear acceleration and orientation during initial testing.

\begin{equation}
  F_{speed} = \frac{\lVert P_{end} - P_{start} \rVert}{time_{end} - time_{start}}
  \label{eq.fitSpeed}
\end{equation}

\begin{equation}
  F_{stability} = -\sum\limits_{i}^{axes}{\Big(\alpha * std(Acc_{i}) + std(Ang_{i})\Big)}
  \label{eq.fitStab}
\end{equation}

The two fitness objectives are both needed.
Evolving only for speed incentivizes solutions that are at the brink of falling, and evolving only for stability incentivizes solutions that move as little as possible.
Combining the two objectives results in the ability to choose robust trade-offs over a range of different speed-stability combinations.

Our genetic algorithm uses Gaussian mutation with a probability of 1 and a sigma of 1/6, with no recombination. 
This has been shown to give a good balance between exploration and exploitation in previous experiments~\citep{tonnesfn_ices16}.
To prevent values accumulating around the upper and lower bounds, we continue mutation back towards the middle of the range when we hit a maximum or minimum value.

Before evaluating a new individual, the length of the legs is changed, and the feet repositioned one at a time to the start pose of the new gait, to remove any measurement contamination from previous individuals.
Each evaluation was completed when either the robot had walked one meter forwards or 10 seconds had elapsed.
Previous experiments have shown this to give a reasonable estimation of the actual performance of both fast and slow gaits, regardless of stability.
The robot was then manually moved back to the starting position, before the next individual was evaluated.
The operator had a remote control with the ability to pause, retry, or discard individuals if anything unexpected happened.
This is done to not penalize individuals for effects out of the robots control, for instance the operator tripping.
Servo temperatures were monitored, and the robot was left to cool down when any servo exceeded 60$^{\circ}$C.
The evaluation time for a single individual was approximately 30 seconds, including setup, cooldown and other manual intervention, as well as the actual walking.

Performing evolution on a real-world robot severely limits the number of evaluations attainable for each run of the algorithm.
The larger the number of evaluations, the higher the probability of damage to the robot that could potentially skew the results.
Based on previous successful experimentation~\citep{tonnesfn_gecco18}, we run 256 evaluations, encompassing 32 generations of 8 individuals.
These parameters promote convergence, without wasting an excessive amount of evaluations at the end of the runs on minor improvements.

\subsection{Experimental Environments}

\begin{figure}
    \centering
    \hspace{7mm} \textsf{Hard} \hspace{55mm} \textsf{Soft}\\\vspace{1mm}
    \begin{sideways}\hspace{15mm}\textsf{Fine}\end{sideways}\hspace{1mm}
    \subfloat{%
    \begin{overpic}[width=0.43\textwidth]{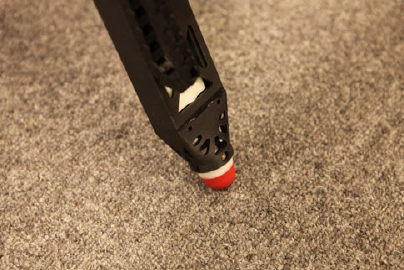}\put (27,4) {%
    \begin{tcolorbox}[colback=white, width=23mm, height=1em+1.5mm, arc=0mm, boxsep=-1.5mm, boxrule=0.1mm]\vspace{0.5mm}\centering \textsf{Surface A}\end{tcolorbox}
    }\end{overpic}%
    \label{fig.surface_basic}}%
    \hspace{1mm}
    \subfloat{%
    \begin{overpic}[width=0.43\textwidth]{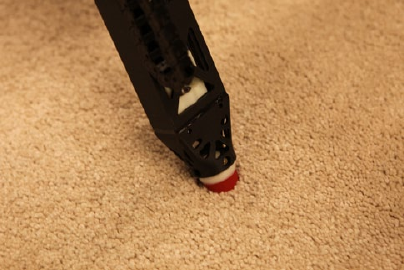}\put (27,4) {%
    \begin{tcolorbox}[colback=white, width=23mm, height=1em+1.5mm, arc=0mm, boxsep=-1.5mm, boxrule=0.1mm]\vspace{0.5mm}\centering \textsf{Surface B}\end{tcolorbox}
    }\end{overpic}%
    \label{fig.surface_fluffy}}%
    \hfil\\\vspace{-2.4mm}
    \begin{sideways}\hspace{13.5mm}\textsf{Coarse}\end{sideways}\hspace{1mm}
    \subfloat{%
    \begin{overpic}[width=0.43\textwidth]{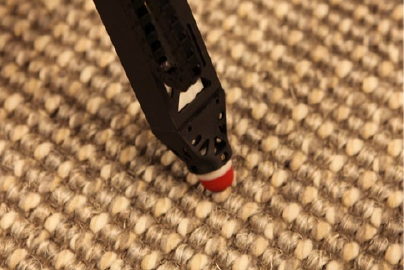}\put (27,4) {%
    \begin{tcolorbox}[colback=white, width=23mm, height=1em+1.5mm, arc=0mm, boxsep=-1.5mm, boxrule=0.1mm]\vspace{0.5mm}\centering \textsf{Surface C}\end{tcolorbox}
    }\end{overpic}%
    \label{fig.surface_rough}}%
    \hspace{1mm}
    \vspace{1mm}
    \subfloat{%
    \begin{overpic}[width=0.43\textwidth]{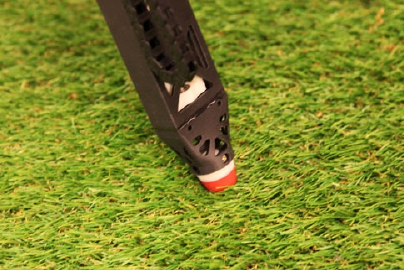}\put (27,4) {%
    \begin{tcolorbox}[colback=white, width=23mm, height=1em+1.5mm, arc=0mm, boxsep=-1.5mm, boxrule=0.1mm]\vspace{0.5mm}\centering \textsf{Surface D}\end{tcolorbox}
    }\end{overpic}%
    \label{fig-surface_grass}}
    \caption{The four different carpets used to approximate real-world terrains with different characteristics.}
    \label{fig.surfaces}
\end{figure}

The purpose of our experimentation is to assess how different environments affect the evolution of different individuals, looking at both performance and behavior.
Robots operate in an increasing number of terrains with varying roughness, slope, discontinuity, and hardness characteristics~\citep{Howard2001}.
We select four surfaces that can be characterized by the two qualitative features, shown as different dimensions in Figure \ref{fig.surfaces}: hardness (soft and hard) and roughness (coarse-textured and fine-textured). 
In our experiments, we consider hardness to be the primary feature (used to train two distinct populations) and roughness to be a secondary feature (used to test how these populations generalize to new environments).
Surface A serves as a baseline and is a hard, fine-textured surface with high friction.
Surface B is also fine-textured, but very soft, so the robot's legs easily sink into it.
Surface C is hard, with large knots that give it a coarse texture.
Surface D is soft, with individual strands in multiple directions giving it a coarser, non-uniform texture.
These surfaces reflect features found in natural terrains like concrete, sand, hard-packed soil, and grass, respectively.

\section{Experiments and Results}

We conducted two experiments.  We first ran 10 full evolutionary runs on the fine-textured hard (A) and fine-textured soft (B) surfaces. We then randomly selected 12 individuals from the final Pareto front and re-evaluated them on all four surfaces to observe how the evolved individual's performance compares on the previously unseen coarse-textured hard (C) and coarse-textured soft (D) surfaces.

\subsection*{Experiment 1 - Evolutionary Runs}
In this experiment, we ran five evolutionary runs on surface A and five on surface B, producing two Pareto fronts of individuals evolved for the different surface hardnesses. 
We alternated runs on the two surfaces to make sure gradual wear-and-tear of the robot would affect results from the two surfaces equally.

\begin{figure}
  \centering
  \includegraphics{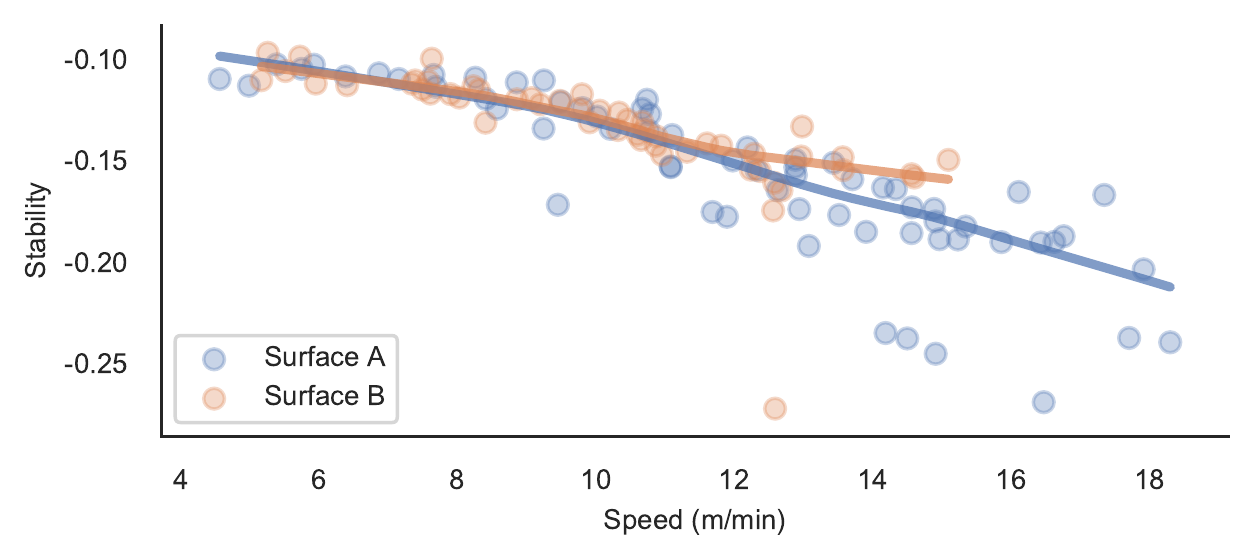}
  \caption{Performance of all individuals from the Pareto fronts of the evolutionary runs on two different surfaces. The lines are a result of locally weighted linear regression using a nonparametric lowess model with default Seaborn parameters.}
  \label{fig.rw01_pareto}
\end{figure}
\begin{figure}
  \centering
  \includegraphics{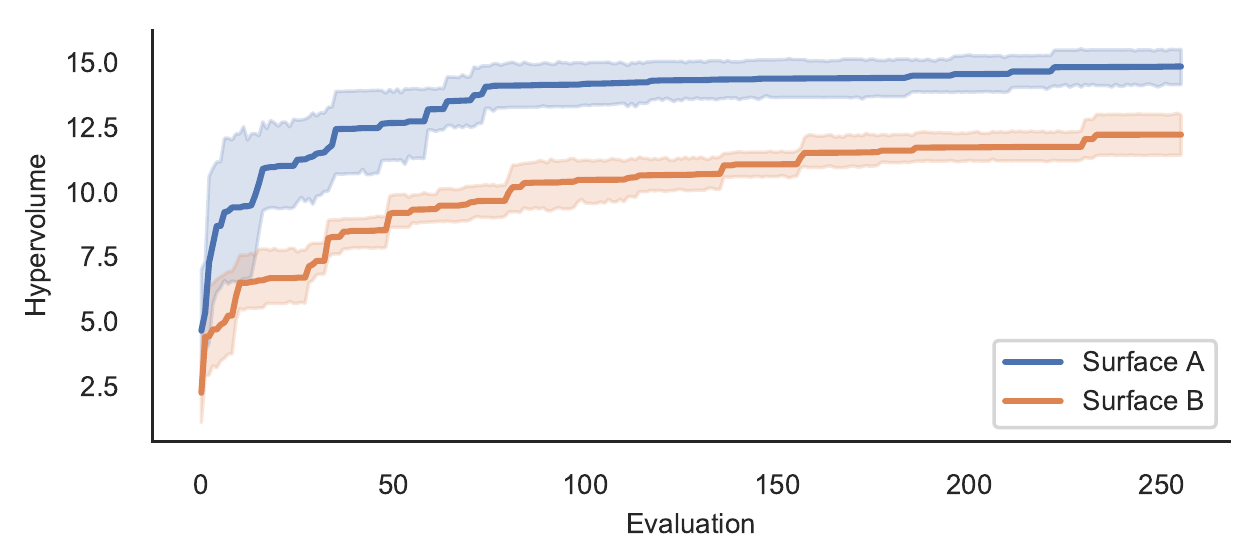}
  \caption{Mean hypervolume of the Pareto fronts from the evolutionary runs on each surface. The hypervolume is calculated as the size of the dominated area under the Pareto front, to a minimum speed of 0, and a minimum stability of -1. The shaded areas show the 95\% confidence interval for the mean value.}
  \label{fig.rw01_convergence}
\end{figure}

Fig.~\ref{fig.rw01_pareto} shows the resulting Pareto fronts from all the evolutionary runs.
The individuals achieved similar performance for speeds below about 10m/min, and higher fitness on the softer carpet (surface B) for speeds between about 10m/min and 15m/min.
Speeds above 15m/min were only seen on the hard carpet (surface A).
For each surface, four out of five evolutionary runs had individuals that were part of the final global Pareto front.

Fig.~\ref{fig.rw01_convergence} shows the development of the mean hypervolume of the Pareto fronts as the search progresses.
The search appears to have converged on the hard carpet (surface A) after less than 100 evaluations. 
The search on the soft carpet (surface B), however, shows slight improvements up to the end of the evolutionary runs.

\begin{figure}
  \includegraphics{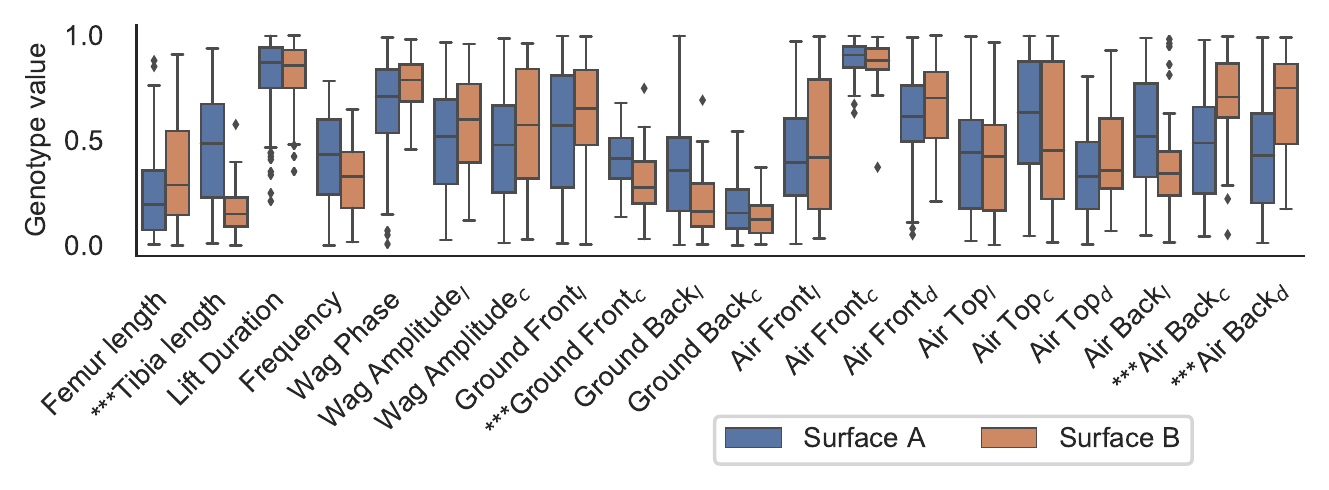}
  \caption{The genotypic values from all individuals on the final Pareto fronts from all evolutionary runs, grouped by the surface they were evolved on. Control point parameters are denoted with the axis they belong to, with $l$ for lateral, $c$ for cranial, and $d$ for dorsal directions. ***Statistically significant differences.}
  \label{fig.rw01_genotypes_boxplot}
\end{figure}

Fig.~\ref{fig.rw01_genotypes_boxplot} shows the parameter values of the individuals in the Pareto front for all evolutionary runs on each surface. 
We ran a two-sided Mann-Whitney U test on each parameter (n1 = 65, n2 = 52, $p < 0.01$), with a Holm-Bonferroni p-value correction, to see if the individuals found with the evolutionary search had statistically significant differences in some of their parameters for the different surfaces.
For morphology, the tibia length showed significant differences (U = 2733).
For the controller, we found statistically significant differences for the cranial position of the ground front control point (U = 2478), and the cranial (U = 783) and dorsal position (U = 925) of the air back control point.
Note that the parameter values seen are from the full resulting Pareto front, including a range of different trade-offs between speed and stability.

\begin{figure}[h]
    \centering
    \subfloat{\includegraphics[width=0.4\textwidth]{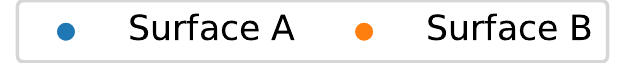}}\\
    \setcounter{subfigure}{0}
    \subfloat[Back air cranial position (mm)]{\includegraphics{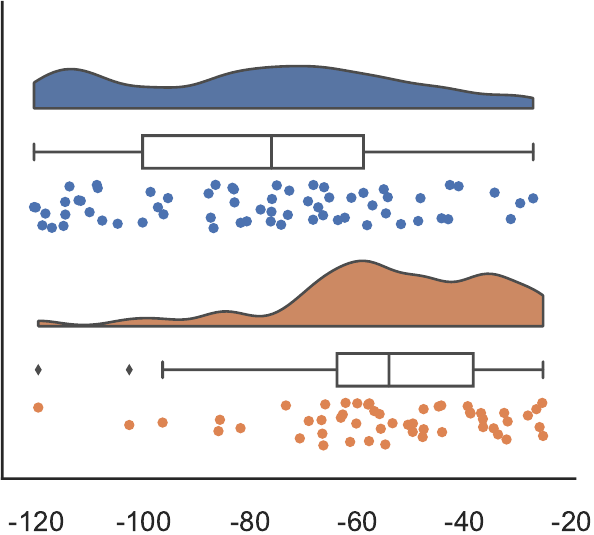}%
    \label{fig.rw01_phenotype_raincloud_p4y}}%
    \hspace{5mm}
    \subfloat[Back air dorsal position (mm)]{\includegraphics{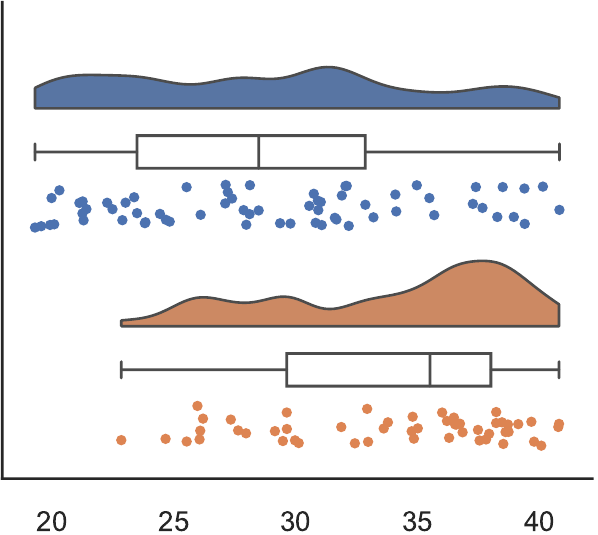}%
    \label{fig.rw01_phenotype_raincloud_p4z}}%
    \\
    \subfloat[Femur length (mm)]{\includegraphics{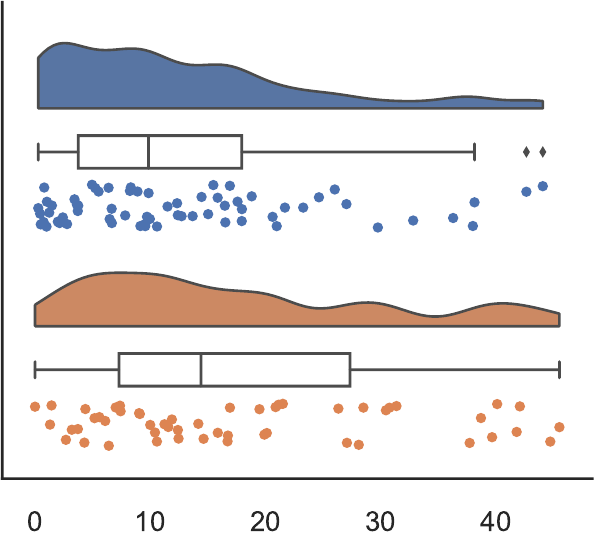}%
    \label{fig.rw01_phenotype_raincloud_femurLength}}%
    \hspace{5mm}
    \subfloat[Tibia length (mm)]{\includegraphics{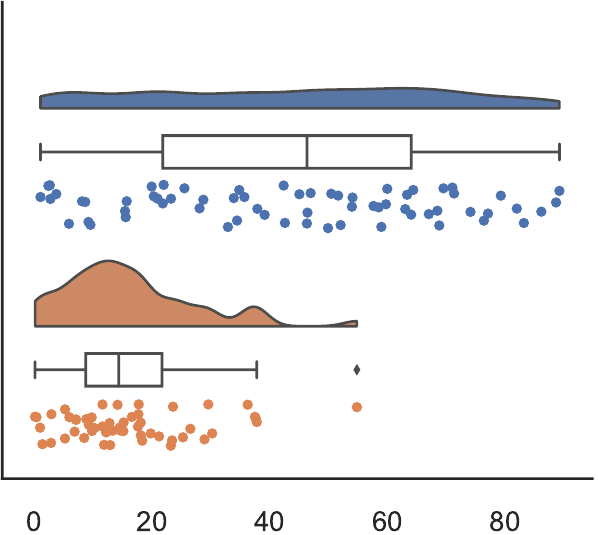}%
    \label{fig.rw01_phenotype_raincloud_tibiaLength}}%

    \caption{Rain cloud plots showing the distributions for two selected control parameters (a-b) and both morphology parameters (c-d).}
    \label{fig.rw01_phenotype_control}
    \vspace{-1mm}
\end{figure}

Fig.~\ref{fig.rw01_phenotype_control} shows the details of two selected parameters from control and both parameters from morphology.
We see little difference in the length of the femur (Fig.~\ref{fig.rw01_phenotype_raincloud_femurLength}.), but the tibia has a clear difference between the two surfaces (Fig.~\ref{fig.rw01_phenotype_raincloud_tibiaLength}.).
On the hard carpet (surface A), we see a very uniform distribution across the whole range.
However, for the soft carpet (surface B), we only see one single individual with a tibia length above 40\% of the available length.

\begin{figure}
  \centering
  \includegraphics[width=0.7\textwidth]{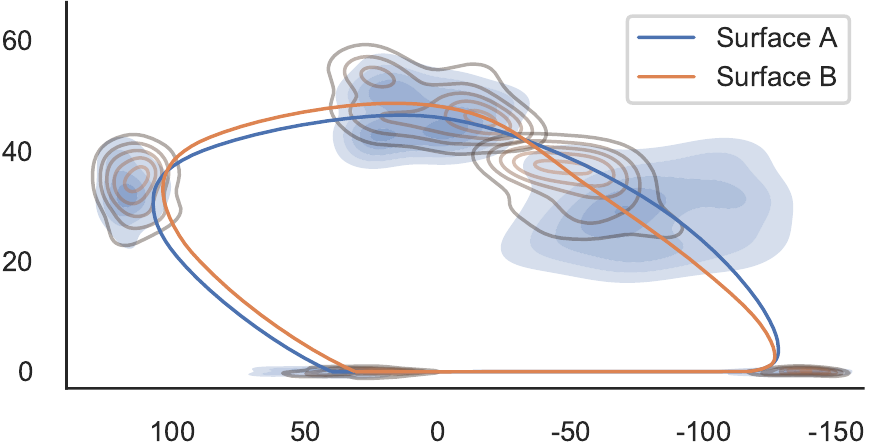}
  \caption{The solid lines show the mean trajectory splines of all Pareto-optimal individuals from each evolutionary run. The shaded areas show the kernel density estimates for the spline control points. The plot is seen from the side of the robot, with the front of the robot to the left of the figure.}
  \label{fig.rw01_splineControl_combined}
\end{figure}

To better visualize the differences in the evolved controller parameters, Fig~\ref{fig.rw01_splineControl_combined} shows kernel density estimates for the control points, as well as the mean controller splines.
The mean splines were found by averaging all positions for each time step in the gait period.
The kernel density estimates are shown as shaded areas and were calculated using a Gaussian kernel with Scott's rule to determine kernel size.
The figure shows small differences in the two points on the ground, as well as the front and top points in the air, but there are more substantial differences in the back air point.

\vspace{3mm}

From these results, we see statistically significant effects on both control and morphology due to the impact of evolving on different surfaces. 

\subsection*{Experiment 2 - Verification}
In this experiment, we verify the results from the evolutionary runs by re-evaluating a subset of the individuals multiple times on the surface they were evolved on, as well as observing the performance on unseen surfaces.
We selected six random individuals from the resulting Pareto fronts of the evolutionary runs for each surface.
We then tested these twelve individuals 20 times each on all four surfaces, for a total of 80 new evaluations per individual.

\begin{figure}
    \centering
    \subfloat{\includegraphics{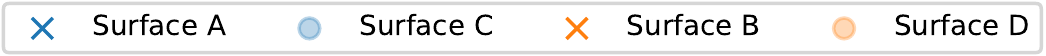}%
    \label{fig.rw02_charles_legend}}\vspace{-2mm}\\%
    \subfloat{\includegraphics{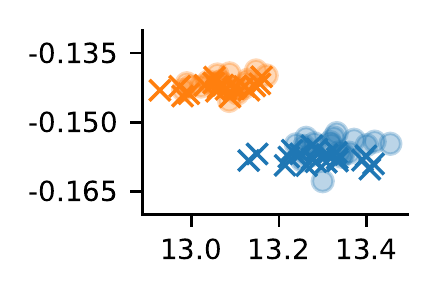}%
    \label{fig.rw02_charles_0}}%
    \subfloat{\includegraphics{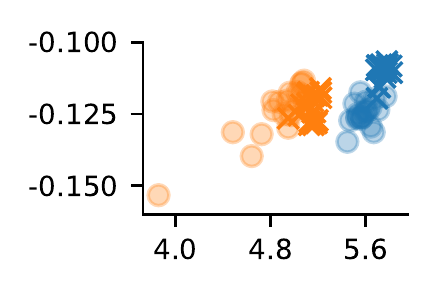}%
    \label{fig.rw02_charles_1}}%
    \subfloat{\includegraphics{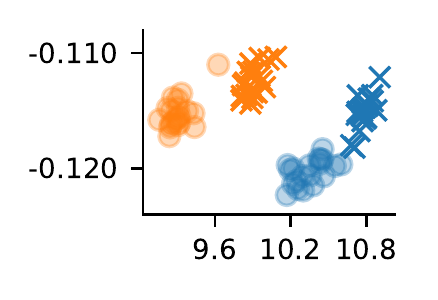}%
    \label{fig.rw02_charles_2}}\vspace{-5mm}\\%
    \subfloat{\includegraphics{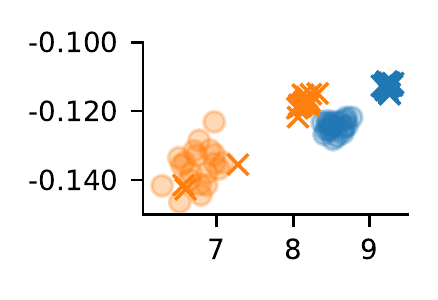}%
    \label{fig.rw02_charles_3}}%
    \subfloat{\includegraphics{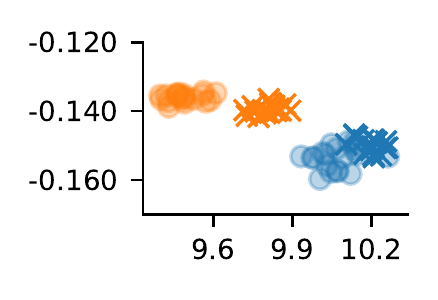}%
    \label{fig.rw02_charles_4}}%
    \subfloat{\includegraphics{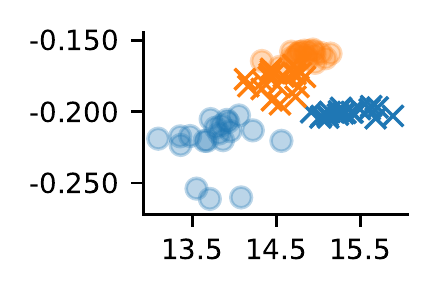}%
    \label{fig.rw02_charles_5}}\vspace{-5mm}\\%
    \subfloat{\includegraphics{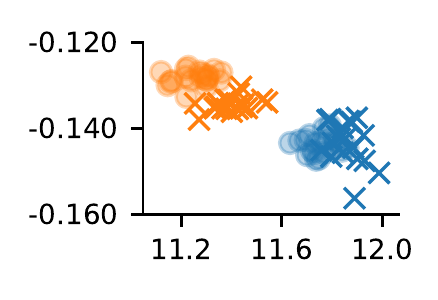}%
    \label{fig.rw02_charles_6}}%
    \subfloat{\includegraphics{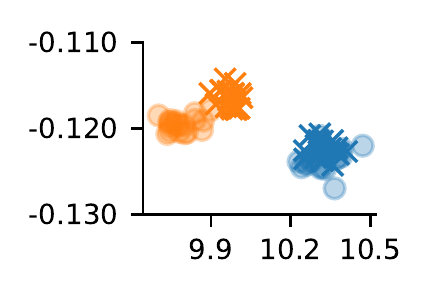}%
    \label{fig.rw02_charles_7}}%
    \subfloat{\includegraphics{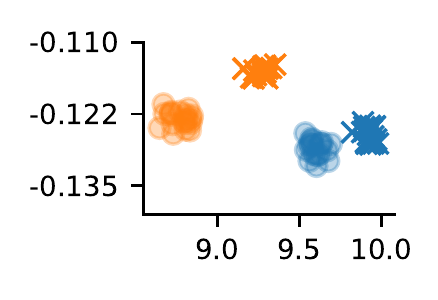}%
    \label{fig.rw02_charles_8}}\vspace{-5mm}\\%
    \subfloat{\includegraphics{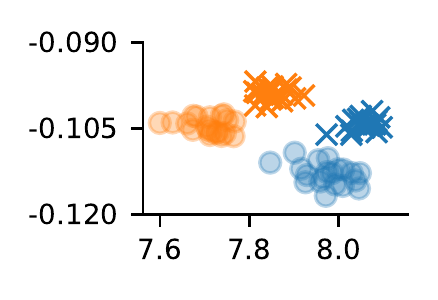}%
    \label{fig.rw02_charles_9}}%
    \subfloat{\includegraphics{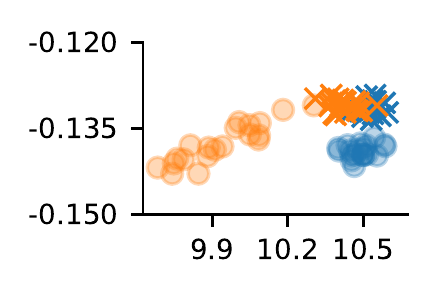}%
    \label{fig.rw02_charles_10}}%
    \subfloat{\includegraphics{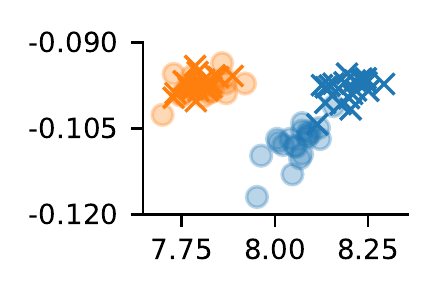}%
    \label{fig.rw02_charles_11}}%
    \caption{Performance measurements for all re-evaluated individuals, with the same axes as Fig. \ref{fig.rw01_pareto}. Each plot contains the 80 evaluations, distributed across the four surfaces, of one individual. The individuals in the two top rows were initially evolved for surface A, while the bottom two were evolved for surface B. Note that the scaling is different for each plot to best show relative differences in performance.}
    \label{fig.rw02_charles}
\end{figure}

Fig.~\ref{fig.rw02_charles} shows the performance of all the re-evaluated individuals.
The performance on surface A and C (both in shades of blue), and B and D (both in shades of orange) are grouped in most of these plots.
This suggests that the performance of individuals tends to be grouped when hardness is similar, our primary feature.

To quantify these differences in performance between the different surfaces, we first did min-max normalization of speed and stability. 
We then measured the Euclidean distance between the mean performance on the different surfaces, for each individual.
The mean results over all individuals are shown in Table~\ref{table.rw02_distanceMatrix}.
This distance matrix shows us the differences in performance between the different surfaces; smaller values show a more similar performance.
We see that the two combinations of surfaces with the smallest difference in performance are surfaces A and C, and B and D.
We also see that performance on surface B is closer to both A and C than performance on surface D is.

From these results, we see that individuals display more similar performance on new surfaces that are closer to the surface they were evolved on.

\begin{table}
  \caption{Distance matrix showing the mean normalized Euclidean distance between evaluations on different surfaces.}
  \label{table.rw02_distanceMatrix}
  \centering
  \begin{tabular}{l l l l}
                                   & Surface A     & Surface C & Surface B             \\ \cline{2-4} 
    \multicolumn{1}{l|}{Surface D} & .108          & .106      & \textbf{.048} \\
    \multicolumn{1}{l|}{Surface B} & .073          & .082      & \textbf{}     \\
    \multicolumn{1}{l|}{Surface C} & \textbf{.053} &           &              
  \end{tabular}
  \vspace{-2mm}
\end{table}

\FloatBarrier

\section{Discussion}

The results in Experiment 1 showed statistically significant effects on both control and morphology due to the impact of evolving on different surfaces. 
This directly supports our first hypothesis (H1) and suggests that there is a benefit of optimizing morphology for different surfaces.
Moreover, the experiment shows that the evolutionary search was able to find suitable individuals adapted to the real-world characteristics of the surfaces, using only real-world evaluations.

The results of Experiment 2 showed that individuals display similar performance on new surfaces that are closer to the surface of their evolution.
This supports the second hypothesis (H2) and suggests that the results apply to qualitatively similar surfaces.

Overall, these two experiments suggest that an archive of different individuals, evolved for a range of different terrains in an offline fashion, could be effectively utilized on new terrains by selecting previously found individuals from similar surfaces in the archive.
The archived individuals could then be further improved by local search or other techniques if needed.

\subsection{Interpretation of the evolved solutions}

We believe that the difference in maximum achieved speed between the two surfaces, as seen in Fig.~\ref{fig.rw01_pareto}, is related to the difference in leg length seen in Fig.~\ref{fig.rw01_phenotype_control}c/d.
The morphologies found for surface A have longer legs, and are thus able to achieve higher leg speeds for the same joint velocity.
This also fits well with the qualitative estimate of walking difficulty, as softer surfaces typically require more power and torque to walk in.
Between about 10m/min and 15m/min, however, individuals evolved on surface B outperforms individuals evolved for surface A.
This is most likely because the search process inherently spends more effort on improving individuals at the ends of the Pareto front, aided by the crowding distance metric in the NSGA-II algorithm.
Small performance increases in this area for surface B will result in the individuals being kept. For surface A, this effort might instead be spent on trying to find faster individuals, which is inherently more difficult, and requires considerable effort.

Fig. \ref{fig.rw01_genotypes_boxplot} shows the differences between the individuals from the Pareto fronts on the two different surfaces.
There is a very large variance in most of the parameters, but this is not surprising since the groups we are comparing contains the whole range of individuals from the Pareto front.
One would expect less variation if comparing individuals evolved for only a single objective. 
Still, we can observe some clear differences between the sets of individuals evolved for different surfaces.
One might think that the parts of the search space that is not populated by any individuals would point to unusable values that should be removed as to make the search space smaller and thus the search more effective. However, these areas might be necessary for other environments or tasks than what we have used here.

One of the most substantial differences in the leg trajectories can be seen where the leg lifts from the ground, at the back of the step (to the right in Fig.~\ref{fig.rw01_splineControl_combined}).
The leg trajectory takes a sharper turn with a more direct path on the soft surface (surface B), while the movement is more rounded on the hard surface (surface A).
The legs of the robot sink into the soft surfaces, which means that there is a larger difference between commanded and actual movement close to the ground than for the harder surfaces.
This might explain the sharper transition of the trajectory for the soft surface (surface B), as the softness of the terrain provides inherent dampening of the movement, and the trajectory can then be sharper than harder surfaces without this.

\subsection{Reflections on real-world evolution}

Performing evolutionary optimization on real-world robots involves a trade-off between search effectiveness on one side and platform performance degradation, on the other side.
While the number of evaluations in our runs were relatively small compared to typical evolutionary experiments using, e.g., simulated robots, the results showed reasonable convergence and performance to support our hypotheses.
However, the approach could benefit from hybridization with other approaches, e.g., using physics simulation as a starting point~\citep{Zagal2007,collins2018towards} or surrogate models~\citep{gaier17, 8202137} to improve the data-efficiency in the search process.

We have designed a platform with self-modifying morphology, enabling practical real-world evolution of morphology and control, and thus exploiting environmental features hard to model in simulations. 
We showed this for internal conditions, namely actuator power, in \citep{tonnesfn_gecco18}, and external conditions, like surface roughness, in this paper.
While the designed platform optimizes its morphology, it comes at the cost of additional hardware required for the mechanism, adding significantly to the total mass of the system. Future robots featuring new materials and mechanisms may allow morphological reconfiguration with less overhead, which might make it even more applicable to real-world problems.

The fact that individuals from four out of five of the evolutionary runs on each surface are part of the global Pareto fronts shows that we did not have substantial changes to the robot or surfaces during our experiments.
Gradual damage to the robot or changes to the surface during operation might result in individuals from early evaluations getting better performance.
However, the fact that both the first and last evolutionary run on each surface resulted in globally Pareto-optimal individuals suggests this is not an issue in our experiments.

\subsection{Future work}

There are many avenues for future work arising from this paper.
When we compare how the individuals perform on different surfaces, we mainly look at the difference in performance.
Analyzing the behavior instead would allow a more thorough investigation into the underlying effects, but would require more effort both during the experiments and after. 
In the same vein, looking at the actual movement of the legs instead of the target position of the actuators could yield new insight into the effects different terrains have on the robot.
Bringing simulation or surrogate models into the mix could allow longer evolutionary runs with better convergence at fewer physical evaluations in the real world.
There is also a range of other optimization techniques that could be done instead of or in combination with evolutionary search.

\section{Conclusion}

In this paper, we showed that evolutionary optimization on a real-world legged robot adapts both morphology and control to different external environments, suggesting that such capabilities could be a key feature of future adaptive robots.

First, we investigated what effects different ground surfaces have on evolved individuals.
We observed that there are statistically significant differences in both the control and morphology of the evolved individuals, showing that the search can adapt the robot to different physical environments.
We also investigated generalization by testing individuals on previously unseen surfaces, where we observed lower performance differences on surfaces qualitatively similar to the ones they were initially evolved on.

Our results suggest that real-world evolutionary optimization is a suitable technique for adapting both the body and control of a physical legged robot to new physical environments.
We also demonstrated how evolved individuals transfer better to qualitatively similar surfaces.
This means that for many applications, one might not need to do a continuous adaptation to new environments during operation, but leverage an archive of different morphology-controller pair generated in safer, but qualitatively similar, indoor terrains.

Our self-modifying quadruped platform allowed us to perform a large number of real-world evaluations and thus discover solutions exploiting the physical characteristics of different surfaces. 
We think this extension into the real-world domain is a promising approach for the simultaneous optimization of morphology and control for legged robots.
We aim to bring our work out of the lab and into realistic terrains outside, and hope our work inspires others to also take on the exciting challenges faced when doing evolutionary robotics in the real world.

\small

\bibliographystyle{apalike}
\bibliography{bibliography}

\end{document}